\documentclass[letterpaper, 10 pt, conference]{ieeeconf}
\IEEEoverridecommandlockouts

\usepackage{bm}
\usepackage{amsmath} % assumes amsmath package installed
\usepackage{amssymb}  % assumes amsmath package installed
\usepackage{amsfonts}

\usepackage{subfigure}
\usepackage{graphicx}
\usepackage{hyperref}

\usepackage{booktabs}
\usepackage{multirow}
\usepackage{tablefootnote}
\usepackage{threeparttable}

\usepackage{mathtools}
\usepackage{arydshln}
\usepackage{xcolor}
\usepackage{pifont}% http://ctan.org/pkg/pifont
\newcommand{\cmark}{\ding{51}}%
\newcommand{\xmark}{\ding{55}}%
\newcommand{\etal}{\mbox{\emph{et al.}}}
\definecolor{aureolin}{rgb}{0.99, 0.9, 0.0}
\definecolor{caribbeangreen}{rgb}{0.0, 0.8, 0.6}

\title{\LARGE \bf
MoPA: Multi-Modal Prior Aided Domain Adaptation\\ for 3D Semantic Segmentation
}

\author{
Haozhi Cao\textsuperscript{1}, 
Yuecong Xu\textsuperscript{2}, 
Jianfei Yang\textsuperscript{2}, 
Pengyu Yin\textsuperscript{1}, 
Shenghai Yuan\textsuperscript{1}, 
Lihua Xie\textsuperscript{1}, \emph{Fellow, IEEE}
\thanks{\textsuperscript{1}Authors are with the Centre for Advanced Robotics Technology Innovation (CARTIN),  School of Electrical and Electronic Engineering, Nanyang Technological University, Singapore.
{\tt\small \{haozhi002, pengyu001, shyuan, elhxie\}@ntu.edu.sg}}%
\thanks{\textsuperscript{2}Authors are with the School of Electrical and Electronic Engineering, Nanyang Technological University, Singapore.
{\tt\small \{xuyu0014, yang0478\}@ntu.edu.sg}}%
\thanks{This research is supported by the National Research Foundation, Singapore under its Medium Sized Center for Advanced Robotics Technology Innovation.}
}
\begin{document}

\hbadness=2000000000
\vbadness=2000000000
\hfuzz=100pt

\maketitle
\thispagestyle{empty}
\pagestyle{empty}

%%%%%%%%%%%%%%%%%%%%%%%%%%%%%%%%%%%%%%%%%%%%%%%%%%%%%%%%%%%%%%%%%%%%%%%%%%%%%%%%
\begin{abstract}
Multi-modal unsupervised domain adaptation (MM-UDA) for 3D semantic segmentation is a practical solution to embed semantic understanding in autonomous systems without expensive point-wise annotations. While previous MM-UDA methods can achieve overall improvement, they suffer from significant class-imbalanced performance, restricting their adoption in real applications. This imbalanced performance is mainly caused by: 1) self-training with imbalanced data and 2) the lack of pixel-wise 2D supervision signals. In this work, we propose Multi-modal Prior Aided (MoPA) domain adaptation to improve the performance of rare objects. Specifically, we develop Valid Ground-based Insertion (VGI) to rectify the imbalance supervision signals by inserting prior rare objects collected from the wild while avoiding introducing artificial artifacts that lead to trivial solutions. Meanwhile, our SAM consistency loss leverages the 2D prior semantic masks from SAM as pixel-wise supervision signals to encourage consistent predictions for each object in the semantic mask. The knowledge learned from modal-specific prior is then shared across modalities to achieve better rare object segmentation. Extensive experiments show that our method achieves state-of-the-art performance on the challenging MM-UDA benchmark. Code will be available at \href{https://github.com/AronCao49/MoPA}{https://github.com/AronCao49/MoPA}.
\end{abstract}

%%%%%%%%%%%%%%%%%%%%%%%%%%%%%%%%%%%%%%%%%%%%%%%%%%%%%%%%%%%%%%%%%%%%%%%%%%%%%%%%
\section{Introduction}
3D scene understanding plays an important role in various robotics tasks, such as localization~\cite{nguyen2018robust, chen2019suma++, ji2021towards, yin2023segregator} and path planning~\cite{bartolomei2020perception, stache2023adaptive, liu2023non}. For intelligent robotics (e.g., autonomous driving and navigation), 3D semantic information about the surrounding environments is a must and therefore has attracted increasing attention from the community. While previous methods~\cite{tang2020searching, zhu2021cylindrical, yan20222dpass} have achieved outstanding performance based on fully supervised learning, they require expensive annotations and suffer from performance degeneracy when used in unseen environments. 

\begin{figure}[t]
    \centering
    \includegraphics[width=.75\linewidth]{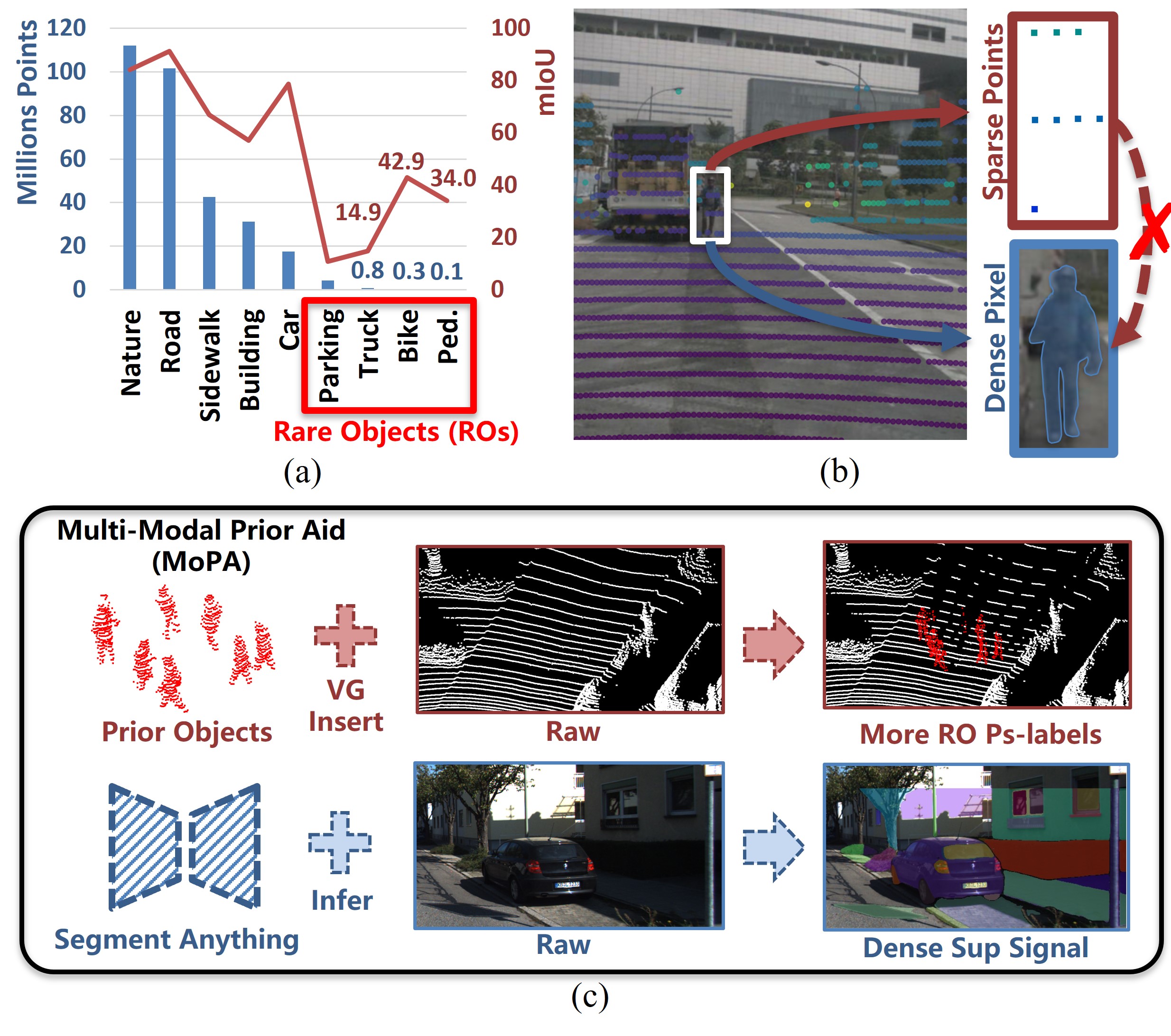}
    \vspace*{-8pt}
    \caption{Illustration of factors that lead to unsatisfactory Rare Objects (ROs) performances (a), (b) and our methods (c). (a) shows the unsatisfactory performances of  ROs in a long-tailed dataset (SemanticKITTI~\cite{behley2019semantickitti}). (b) illustrates the limited supervision signals provided by point-wise pseudo-labels (only 8 points) to guide the 2D networks to identify the pedestrian in the dense image. (c) demonstrates the overall structure of our proposed MoPA, which introduces more prior ROs through VGI and dense supervision signals guided by Segment Anything Model (SAM)~\cite{kirillov2023segment}}
    \label{Fig:Intro}
    \vspace{-6pt}
\end{figure}

To alleviate these limitations, Unsupervised Domain Adaptation (UDA)~\cite{jiang2021lidarnet,rochan2022unsupervised,tsai2023viewer} has been proposed to leverage the knowledge from the labeled source domain (e.g., public datasets with delicate annotations) to the unlabeled target domain (e.g., specific application environments) for 3D semantic segmentation. With an increasing number of multi-modal datasets~\cite{behley2019semantickitti, caesar2020nuscenes, sun2020scalability}, Multi-Modal UDA (MM-UDA) has been proposed to leverage the complementary information of 2D images and 3D point clouds for segmentation. As the primary work, xMUDA~\cite{jaritz2020xmuda} proposes to encourage cross-modal prediction consistency. The following works further refine this framework from different perspectives (e.g., diversifying pixel-point corresponding~\cite{peng2021sparse, xing2023cross} and reducing domain discrepancy~\cite{liu2021adversarial, li2022cross}). While achieving noticeable improvement, previous works barely reveal and discuss their class-wise performance. To dig deeper into previous MM-UDA methods, we illustrate the class-wise performance of exemplary xMUDA~\cite{jaritz2020xmuda} in Fig.~\ref{Fig:Intro}(a). One can easily observe that xMUDA suffers from serious class-imbalanced performance, whose main bottleneck lies in the object classes (e.g., ``Truck'', ``'Bike'', and ``Pedestrian''). These classes, however, usually play a more important role in applications involving human-robot interaction (e.g., trajectory-aware path planning and obstacle avoidance). 

Motivated by this observation, we aim to study how to alleviate the class-imbalanced performance in this work. Most existing MM-UDA methods~\cite{jaritz2020xmuda, peng2021sparse, li2022cross} rely on self-training with pseudo-labels (i.e., confident predictions generated by the network itself) to compensate for the lack of ground-truth supervision signals, whereas this strategy is known to be ineffective for Rare Objects (ROs) in a long-tailed dataset as in Fig.~\ref{Fig:Intro}(a). Without effective rectification for this class-wise imbalance, the pseudo-labels generated are biased toward the dominant classes, resulting in a scarcity of true-positive pseudo-labels for ROs. Moreover, since previous methods adopt a similar one-to-one pixel-point corresponding, the pseudo-labels generated are point-wise instead of pixel-wise. In terms of ROs with small LiDAR cross sections (e.g., ``Bike'' and ``Pedestrian'') as shown in Fig.~\ref{Fig:Intro}(b), the point-wise pseudo-labels are too sparse to supervise the dense pixel-wise predictions (e.g., 7 points vs. hundreds of pixels), resulting in inferior RO performances.

% Ideally, prior knowledge can be leveraged to alleviate the aforementioned limitations. For the lack of reliable pseudo-labels for ROs, considering ROs are mostly movable objects, one can introduce prior ROs with accurate semantic labels in the raw input scan. If the introduction is flawless enough to avoid trivial solutions, the knowledge learned to classify these prior ROs could be propagated to detect the original ROs. On the other hand, the lack of dense supervision signals for the 2D network can be compensated by the prior knowledge from a 2D expert model. 

To this end, we propose an MM-UDA method called \textbf{M}ulti-m\textbf{o}dal \textbf{P}rior \textbf{A}ided (MoPA) that boosts the ROs segmentation performance by addressing the aforementioned limitations as shown in Fig.~\ref{Fig:Intro}(c). To rectify the imbalanced pseudo-labels, MoPA randomly inserts prior ROs with accurate labels collected from the wild (i.e., 3D prior knowledge) in each raw scan. Considering that directly inserting ROs can compromise the raw scan and lead to trivial solutions, Valid Ground-based Insertion (VGI) is developed to ensure the effectiveness of inserted prior ROs through validity checking and ray-tracing-based style translation. To compensate for the lack of dense 2D supervision signals, MoPA leverages the semantic masks from the foundation model Segment Anything Model (SAM)~\cite{kirillov2023segment} as the 2D prior knowledge with reliable pixel-wise supervision signals. Combined with cross-modal learning, the multi-modal prior knowledge can be propagated across modalities to achieve better RO segmentation.

In summary, our contributions are listed as follows:
\begin{itemize}
    \item We propose an MM-UDA method called MoPA, leveraging multi-modal prior to improve RO segmentation.
    \item VGI is developed to introduce more reliable RO pseudo-labels with validity guarantee and style translation. Meanwhile, SAM consistency is proposed to compensate for the lack of 2D dense supervision signals.
    \item Extensive experiments show that our MoPA achieves state-of-the-art performance on the challenging MM-UDA benchmark for 3D semantic segmentation. 
\end{itemize}

\section{Related Work}
\noindent \textbf{Uni-Modal Domain Adaptation for Segmentation.} Previous UDA methods for semantic segmentation can be divided as follows: (i) domain adversarial discriminative approaches~\cite{ganin2015unsupervised,tzeng2017adversarial}, (ii) generative approaches~\cite{huang2018domain,li2019bidirectional}, (iii) self-training approaches~\cite{zou2018unsupervised,zou2019confidence}, and (iv) uncertainty minimization approaches~\cite{vu2019advent, chen2019domain}. In terms of UDA for 3D segmentation, besides those mostly inspired by the 2D UDA methods~\cite{wu2019squeezesegv2, zhao2021epointda}, some previous works develop UDA methods to address some specific point-cloud-only discrepancy (e.g., the difference of point density~\cite{elhadidy2020improved, yi2021complete}). However, these methods are designed for uni-modal input and thus can not fully leverage the multi-modal information.

% \noindent \textbf{Uni-Modal UDA for Segmentation.}
% Previous UDA methods for semantic segmentation can be generally divided into 4 categories: (i) adversarial-based approaches~\cite{ganin2015unsupervised,tzeng2017adversarial,xu2021aligning}, (ii) generative-based approaches~\cite{huang2018domain,li2019bidirectional}, (iii) self-training-based approaches~\cite{zou2018unsupervised,zou2019confidence}, and (iv) uncertainty-based approaches~\cite{vu2019advent, chen2019domain}. While the above methods are mostly developed for 2D images, recent UDA methods have been proposed specifically for 3D point clouds. These include methods inspired by the 2D UDA methods~\cite{wu2019squeezesegv2, zhao2021epointda}, and works that addresss point-cloud-only discrepancy (e.g., the difference of point density~\cite{elhadidy2020improved, yi2021complete}). However, these methods are designed for uni-modal input and therefore can not fully leverage the multi-modal information.

\noindent \textbf{Multi-Modal Domain Adaptation for Segmentation.}
Thanks to the emerging adaptation of multi-modal sensors and multi-modal datasets~\cite{behley2019semantickitti, geyer2020a2d2, caesar2020nuscenes, sun2020scalability}, domain adaptation with multi-modal input~\cite{jaritz2020xmuda, peng2021sparse, liu2021adversarial, li2022cross, xing2023cross, cao2023multi} has attracted increasing attention. xMUDA~\cite{jaritz2020xmuda} proposes the first MM-UDA method by encouraging cross-modal prediction consistency. DsCML~\cite{peng2021sparse} interprets the point-pixel-correspondence in a deformable manner and extends the cross-modal learning to an inter-domain perspective. The following works follow this cross-modal-consistency strategy and further improve the segmentation performance by introducing additional components (e.g., adversarial learning~\cite{liu2021adversarial}, knowledge distillation~\cite{li2022cross}, and contrastive learning~\cite{xing2023cross}). Despite their consistent improvement in overall performance, none of them discuss the imbalanced class-wise performance of ROs, whose detection, however, can play an important role in other downstream tasks such as navigation and autonomous driving. To fulfill this limitation, we specifically study how to improve the inferior performance of ROs for MM-UDA.

% \noindent \textbf{Multi-Modal UDA for Segmentation.}
% Thanks to the emerging adaptation of multi-modal sensors and multi-modal datasets~\cite{behley2019semantickitti, geyer2020a2d2, caesar2020nuscenes, sun2020scalability}, UDA with multi-modal input~\cite{jaritz2020xmuda, peng2021sparse, liu2021adversarial, li2022cross, xing2023cross, cao2023multi} has attracted increasing attention. xMUDA~\cite{jaritz2020xmuda} proposes the first MM-UDA method by encouraging cross-modal prediction consistency. DsCML~\cite{peng2021sparse} interprets the point-pixel-correspondence in a deformable manner and extends the cross-modal learning to an inter-domain perspective. Subsequent works follow such cross-modal-consistency strategy and further improve performances with additional components (e.g., adversarial learning~\cite{liu2021adversarial}, knowledge distillation~\cite{li2022cross}, and contrastive learning~\cite{xing2023cross}). Despite their consistent improvements, none discussed the imbalanced class-wise performance of rare objects (ROs), whose detection can play an important role in other downstream tasks such as navigation and autonomous driving. To overcome this limitation, we specifically study how to improve the inferior performance of ROs in a cross-modal manner during UDA.

\begin{figure*}[t]
    \centering
    \vspace{5pt}
    \includegraphics[width=.68\textwidth]{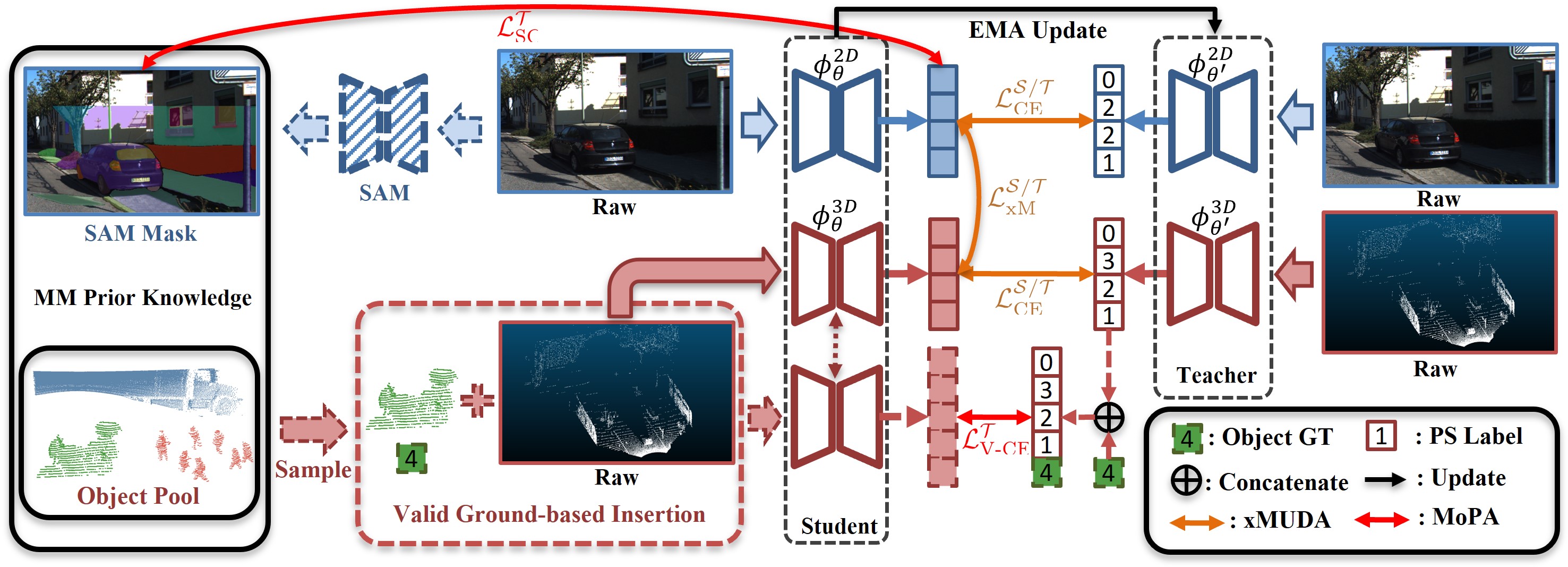}
    \vspace*{-6pt}
    \caption{Overview of our proposed MoPA operating on the target domain. Given a point cloud scan as input, prior objects are randomly sampled from the prior object pool and inserted into the raw input scan (Sec.~\ref{Sec:Method_VGI}) to introduce more ROs with accurate pseudo-labels. Meanwhile, the prior 2D semantic mask is generated from the foundation model SAM~\cite{kirillov2023segment} to serve as our dense 2D supervision signals (Sec.~\ref{Sec:Method_SAM}). Additionally, the EMA module generates online pseudo-labels and facilitates cross-modal learning by introducing cross-modal pseudo-labels. Combined with xMUDA~\cite{jaritz2020xmuda} multi-modal learning pipeline, our MoPA can effectively improve the segmentation performance on ROs by propagating the prior knowledge in a cross-modal manner.}
    \label{Fig:Main_Method}
    \vspace{-10pt}
\end{figure*}

\noindent \textbf{Mixing-based Strategies.}
Sample mixing has been widely adopted in the image domain~\cite{zhang2017mixup, berthelot2019mixmatch, yun2019cutmix}, which has shown its effectiveness in reducing domain discrepancy. Inspired by this idea, some recent works have proposed different mixing strategies for point cloud classification~\cite{chen2020pointmixup, zhang2022pointcutmix} or segmentation~\cite{saltori2022cosmix, hasecke2022fake, kong2023conda}. For segmentation, CosMix~\cite{saltori2022cosmix} concatenates the confident semantic point cloud from the source domain to the target domain and vice versa. Hasecke \etal~\cite{hasecke2022fake} proposes to regenerate point clouds to mitigate the domain gap caused by different sensor setups. By projecting point clouds to range images, ConDA~\cite{kong2023conda} constructs an intermediate domain through the concatenation of range images. While performance improvement can be achieved, they also introduce some artificial artifacts during the sample mixing of point clouds. In this work, we propose a Valid Ground-based Insertion method that aims to improve the RO class-wise performance for MM-UDA without introducing artificial artifacts.

% \noindent \textbf{Mixing-based Strategies.}
% Sample mixing has been widely adopted in the image domain~\cite{zhang2017mixup, berthelot2019mixmatch, yun2019cutmix}, which has shown its effectiveness in mitigating domain discrepancy. Inspired by this idea, recent works have proposed various mixing strategies for point cloud classification~\cite{chen2020pointmixup, zhang2022pointcutmix} and segmentation~\cite{saltori2022cosmix, hasecke2022fake, kong2023conda}. Among them, CosMix~\cite{saltori2022cosmix} concatenates the confident semantic point cloud from the source to the target domain and vice versa. Hasecke \etal~\cite{hasecke2022fake} proposes to regenerate point clouds to mitigate the domain gap caused by different sensor setups. By projecting point clouds to range images, ConDA~\cite{kong2023conda} constructs an intermediate domain through the concatenation of range images. While improving performances, they also introduce artificial artifacts during the sample mixing of point clouds, which we avoided through proposing a Valid Ground-based Insertion.

\section{Methodology}
\subsection{Preliminary and Overview}
In MM-UDA, the labeled source domain is denoted as $\mathcal{S}=\{(\mathbf{x}_{\mathcal{S}, i}^{\textrm{2D}}, \mathbf{x}_{\mathcal{S}, i}^{\textrm{3D}}, \mathbf{y}_{\mathcal{S}, i})\}^{|S|}_{i=1}$, where $\mathbf{x}_{\mathcal{S}, i}^{\textrm{2D}} \in \mathbb{R}^{H\times W\times 3}$ and $\mathbf{x}_{\mathcal{S}, i}^{\textrm{3D}} \in \mathbb{R}^{N\times3}$ denote the 2D RGB image and 3D point cloud (i.e., the channel includes x, y, and z). $\mathbf{y}_{\mathcal{S}, i} \in \mathbb{R}^{N\times C}$ is the one-hot point-wise semantic label, where $i$ and $C$ denotes the sample index and the number of classes, respectively. The goal is to leverage the knowledge learned from $S$ to an unlabeled target domain $\mathcal{T}=\{(\mathbf{x}_{\mathcal{T}, j}^{\textrm{2D}}, \mathbf{x}_{\mathcal{T}, j}^{\textrm{3D}})\}^{|\mathcal{T}|}_{j=1}$. Following previous methods~\cite{jaritz2020xmuda, peng2021sparse}, we introduce modal-specific networks $\phi_{\theta}^{m}(\cdot), m\in\{\textrm{2D},\textrm{3D}\}$ to process 2D and 3D input independently. Given the model-specific prediction $\mathbf{p}^{m}_{\mathcal{D}}=\phi_{\theta}^{m}(\mathbf{x}_{\mathcal{D}}^{m}), \mathcal{D}\in \{\mathcal{S}, \mathcal{T}\}$, the cross-modal prediction $\mathbf{p}^{\textrm{xM}}_{\mathcal{D}}$ is computed as their average following the same point-to-pixel projection protocol as in~\cite{jaritz2020xmuda, peng2021sparse}. In the following sections, the subscript of the sample index is omitted for clarity.

As shown in Fig~\ref{Fig:Intro}(a), a significant bottleneck of current MM-UDA methods is the unsatisfactory performance of ROs due to (i) limited accurate pseudo-labels from RO classes and (ii) the lack of effective RO supervision signals on the 2D image. Motivated by such observation, our MoPA is designed to address the aforementioned gaps by fully leveraging multi-modal prior knowledge. Given an RO pool as 3D prior knowledge, we propose to insert more prior ROs with accurate labels through Valid Ground-based Insertion (VGI). (Sec.~\ref{Sec:Method_VGI}). Meanwhile, we leverage the 2D prior knowledge from foundation models to compensate for the lack of dense ROs supervision signals (Sec.~\ref{Sec:Method_SAM}). Fig.~\ref{Fig:Main_Method} further illustrates the overall pipeline of MoPA (Sec.~\ref{Sec:Method_Overall}).

\subsection{Valid Ground-based Insertion (3D)}\label{Sec:Method_VGI}
Point cloud mixing~\cite{saltori2022cosmix, kong2023conda} has been known as an effective method for domain adaptation, which mitigates the domain discrepancy by constructing an intermediate domain. Different from previous works, our goal is to introduce more prior ROs with accurate pseudo-labels and propagate the knowledge learned from prior ROs to those in the original scan. To achieve this goal, VGI is designed to insert prior ROs in the input scan with validity checking and style translation to avoid trivial solutions. 

\subsubsection{Preparation of the Object Pool} Before insertion, an object pool is collected from the wild. Given an arbitrary scan of the object source, we preserve the points with the semantic labels of interest and utilize DBSCAN~\cite{ester1996density} to cluster the object instances, leading to an object pool denoted as:
\begin{align}
    \mathcal{Q} = \{(\mathbf{x}_{\mathcal{Q}, k}, \mathbf{y}_{\mathcal{Q}, k)}\}^{|\mathcal{Q}|}_{k=1},
\end{align}
where $\mathbf{x}_{\mathcal{Q}, k}$ and $\mathbf{y}_{\mathcal{Q}, k}$ indicate the 3D points and the semantic label of instance $k$.

\subsubsection{Overlap Checking and Grounding}
Given any raw input scan $\mathbf{x}_{\mathcal{T}}^{\textrm{3D}}$, VGI randomly samples object instances from the object pool and inserts them in the target scan. Unlike previous point cloud concatenation methods~\cite{saltori2022cosmix,kong2023conda}, which directly mix point clouds without validity checking, our insertion process follows two main steps to avoid introducing artificial artifacts: (i) voxel-based overlap checking and (ii) object grounding. For voxel-based overlap checking, the valid location candidates are generated as the inserted object center locations that do not compromise the raw input scan. Specifically, given $N_q$ voxelized prior objects sampled from $\mathcal{Q}$ denoted as $\hat{\mathcal{Q}}' = \{\hat{\mathbf{x}}_{q, n}, \hat{\mathbf{y}}_{q, n}\}^{N_q}_{n=1}$, the query of overlap checking $\hat{\mathbf{q}} = [\hat{e}_x, \hat{e}_y, \hat{e}_z]$ is the 3D extent of the minimum axis-aligned bounding box that includes all prior objects $\hat{\mathbf{x}}_{q, n} \in \hat{\mathcal{Q}}'$. Subsequently, the overlap checking is conducted as the grid-search between the voxel grids of the query $\hat{\mathbf{q}}$ and the searching area in the voxelized input scan $\hat{\mathbf{x}}_{\mathcal{T}}^{\textrm{3D}}$:
\begingroup
\allowdisplaybreaks
\begin{align}
    \hat{\mathbf{Q}} \in& \mathbb{R}^{\hat{e_x} \times \hat{e_y} \times \hat{e_z}}, \hat{\mathbf{Q}}_{ijk} = 1, \\
    \hat{\mathbf{K}} \in& \mathbb{R}^{\Delta x \times \Delta y \times \Delta z}, \hat{\mathbf{K}}_{ijk} = 
        \begin{cases}
            1, &\textrm{if occupied by }\hat{\mathbf{x}}_{\mathcal{T}}^{\textrm{3D}}\\
            0, &\textrm{otherwise}
        \end{cases}, \\
    \hat{\mathbf{V}} =& \{(\hat{x}_i, \hat{y}_i, \hat{z}_i) | (\hat{\mathbf{Q}} * \hat{\mathbf{K}})_{\hat{x}_i\hat{y}_i\hat{z}_i} = 0\},\label{Eq:Overlap_Check}
\end{align}
\endgroup
where $\Delta x, \Delta y, \Delta z$ are the corresponding axis-along extents of the searching area pre-defined by its corner coordinates $(x_1, y_1, z_1)$ and $(x_2, y_2, z_2)$ (i.e., $\Delta x=|x_1 - x_2|$, likewise for $\Delta y$ and $\Delta z$). $\hat{\mathbf{Q}}$ and $\hat{\mathbf{K}}$ denote the voxel grids of the query and the searching area, while $*$ represents the 3D convolution computation. Based on the overlap checking results from Eq.~\ref{Eq:Overlap_Check}, the voxelized valid location candidates $\hat{\mathcal{V}}=\{(\hat{x}_i^v, \hat{y}_i^v, \hat{z}_i^v)\}^{|\hat{\mathcal{V}|}}_{i=1}$ are then computed by mapping $\hat{\mathbf{V}}$ back to the input scan coordinate, at each of which the object insertion won't compromise the raw input scan.

\begin{figure}[t]
    \centering
    \vspace{5pt}
    \includegraphics[width=.75\linewidth]{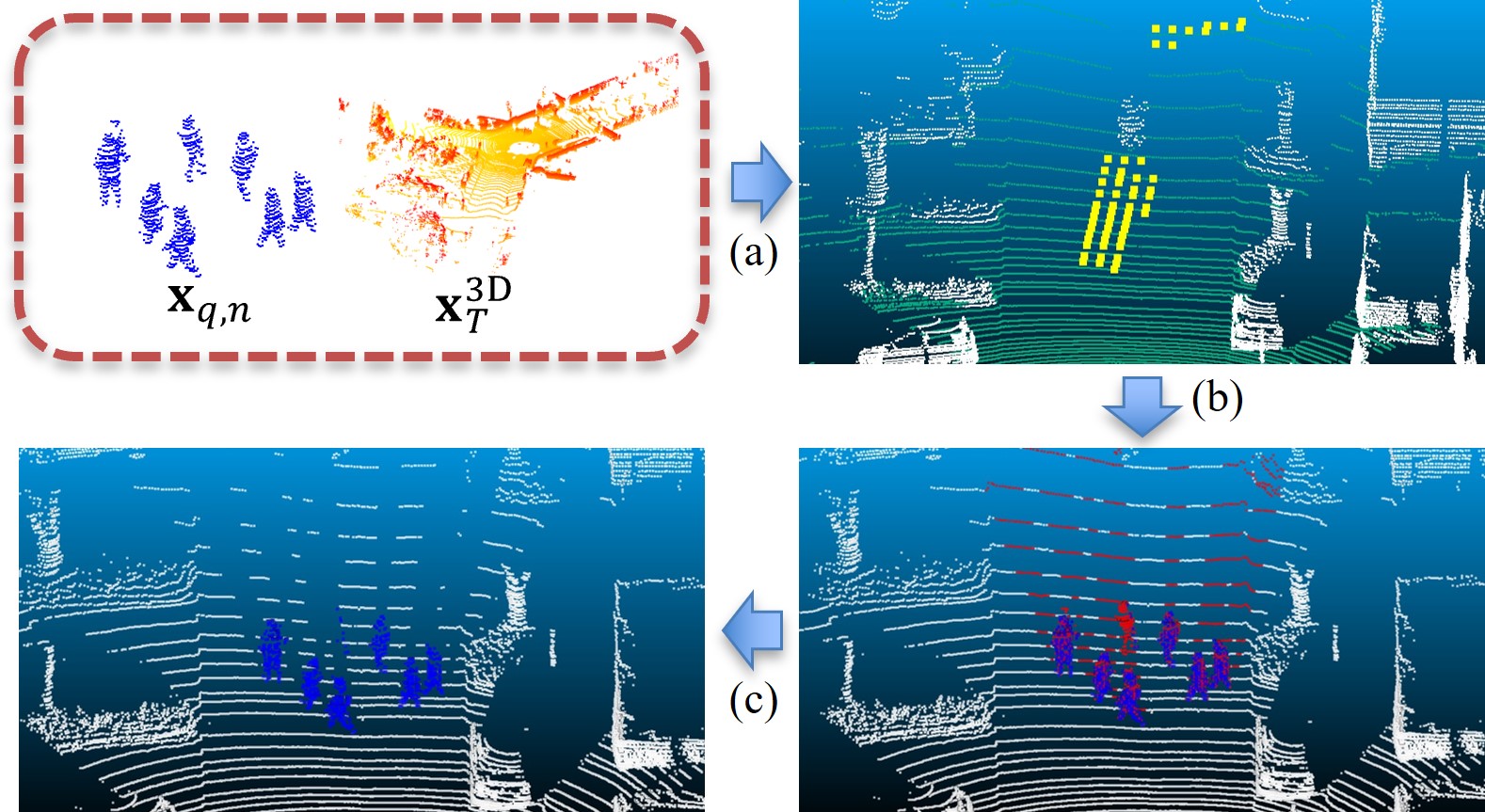}
    \vspace*{-6pt}
    \caption{The detailed steps in VGI. Given a prior RO $\mathbf{x}_{q,n}$ and the raw scan $\mathbf{x}^\textrm{3D}_\mathcal{T}$, the valid insertion locations (highlighted in \textcolor{aureolin}{yellow}) are first computed through (a) overlap checking and grounding, where the \textcolor{caribbeangreen}{green} points are the detected ground. (b) $\mathbf{x}_{q,n}$ is then inserted in the raw scan, whose style is translated by (c) occlusion removal and adaptive downsampling. \textcolor{blue}{Blue} and \textcolor{red}{red} points are the preserved and discarded points in step (c). Best when zoomed in and viewed in color.}
    \smallskip
    \label{Fig:VGI_details}
    \vspace{-10pt}
\end{figure}

Subsequently, the valid location candidates are filtered by only preserving those above the ground as shown in Fig.~\ref{Fig:VGI_details}(a). This is based on the fact that objects are mostly grounded in urban environments. Such a filtration process can simultaneously discard the unreasonable valid location candidates introduced during the grid search (e.g., those behind the building walls). Specifically, given the raw input scan $\hat{\mathbf{x}}_{\mathcal{T}}^{\textrm{3D}}$, we leverage the geometric-based ground detection algorithm Patchwork++~\cite{lee2022patchwork++} as our ground prior thanks to its real-time efficiency, whose detected ground points are similarly voxelized as $\hat{\mathcal{G}}=\{(\hat{x}_i^g, \hat{y}_i^g, \hat{z}_i^g)\}^{|\hat{\mathcal{G}}|}_{i=1}$. Given $\hat{\mathcal{G}}$ and $\hat{\mathcal{V}}$, the valid locations $\hat{\mathcal{V}}_g$ is preserved as follows:
\begingroup
\allowdisplaybreaks
\begin{align}
    \hat{\mathcal{V}}_g = \{&(\hat{x}_i^v, \hat{y}_i^v, \hat{z}_i^v) \in \hat{\mathcal{V}}| \exists j,\nonumber \\
        &(\hat{x}_i^v, \hat{y}_i^v, \hat{z}_i^v - \hat{e}_z/2) = (\hat{x}_j^g, \hat{y}_j^g, \hat{z}_j^g) \in \hat{\mathcal{G}}\}, 
\end{align}
\endgroup
where each insertion location of $\hat{\mathcal{V}}_g$ can be grounded without compromising the raw input scan. $N$ insertion locations, denoted as $\mathcal{C}_{\mathcal{Q}'} = \{(x_i^c, y_i^c, z_i^c)\}^{N_q}_{i=1}$ are then sampled from $\hat{\mathcal{V}}_g$ for each object in $\mathcal{Q}'$. In practice, the altitude of each insertion location is further refined after de-voxelization to ensure the inserted object is grounded in the input scan. 

\subsubsection{Insertion and Style-Translation}
After overlap checking and grounding, the prior object instances are inserted in the raw scan and post-processed for further refinement. Considering the shape of object instances captured by LiDAR sensors is orientation-specific, we preserve the orientation of the prior object instances to further avoid introducing artificial artifacts during the insertion. Given the bounding box center of a prior object instance $\mathbf{x}_{q}$ and its corresponding insertion location denoted as $\mathbf{o}_{q}=(x_{q}, y_{q}, z_{q})$ and $\mathbf{o}_c = (x^c, y^c, z^c), \mathbf{o}_c \in \mathcal{C}_{\mathcal{Q}'}$, the radial translation and z-rotation matrices are formulated as:
\begin{gather}
    \Delta\phi = \phi_c - \phi_{q}, \Delta \rho = \rho_c - \rho_{q}, \\
    \mathbf{t} = 
    \begin{bmatrix}
        \Delta \rho\cos{\phi_{q}} & \Delta \rho\sin{\phi_{q}} & z_c - z_{q}
    \end{bmatrix}, 
    \mathbf{T} = 
    \begin{bmatrix}
        \mathbf{I} & \mathbf{t}^\textrm{T}\\
        \mathbf{0} & 1
    \end{bmatrix},\\
    \mathbf{r} = 
    \begin{bmatrix}
        \cos{\Delta\phi} & -\sin{\Delta\phi}\\
        \sin{\Delta\phi} & \cos{\Delta\phi}
    \end{bmatrix}, 
    \mathbf{R} = 
    \begin{bmatrix}
        \mathbf{r} & \mathbf{0}\\
        \mathbf{0} & \mathbf{I}
    \end{bmatrix},
\end{gather}
where $\phi_{q}$ and $\rho_{q}$ denote the azimuth and radial of the projected point of $\mathbf{o}_{q}$ on the xy-plane (likewise for $\phi_c$ and $\rho_c$). With the radial translation matrix $\mathbf{T}\in\mathbb{R}^{4\times 4}$ and z-rotation matrix $\mathbf{R}\in\mathbb{R}^{4\times 4}$, the transformed prior object $\Tilde{\mathbf{x}}_q=(\mathbf{R}\times\mathbf{T}\times \mathbf{x_q}^{\textrm{T}})^{\textrm{T}}$ is inserted at $\mathbf{o}_c$.

Subsequently, our style-translation stage removes the insertion-induced artificial artifacts that could lead to trivial solutions. This post-processing process adaptively downsamples the point density of $\Tilde{\mathbf{x}}_q$ to a similar level of its surroundings and synthesizes the occlusion introduced by the $\Tilde{\mathbf{x}}_q$. Given the raw input scan $\mathbf{x}_{\mathcal{T}}^{\textrm{3D}}$ and any inserted prior object $\Tilde{\mathbf{x}}_q$, their Range View (RV) image coordinates $\mathbf{x}^{\textrm{RV}}_{\mathcal{T}}\in \mathbb{R}^{N\times2}$ and $\Tilde{\mathbf{x}}^{\textrm{RV}}_{q}\in \mathbb{R}^{M\times2}$ are first computed by the Eq.~(1) in~\cite{milioto2019rangenet++}. For each unique RV coordinate and its corresponding 3D points, the style-translation process acts as an RV-based ray tracing operation, preserving the 3D point with the minimum range while discarding the others as:
% , formulated as:
% Specifically, we project the point clouds of both the raw input scan and the inserted prior objects to the same Range View (RV) image and remove those points whose corresponding RV pixels are occluded by others.
\begingroup
\allowdisplaybreaks
\begin{align}
    (\mathbf{x}^{\textrm{3D}}, \mathbf{x}^{\textrm{RV}}, \Tilde{\mathbf{y}}^{\textrm{3D}}) &= (
    \mathbf{x}_{\mathcal{T}}^{\textrm{3D}} \oplus \Tilde{\mathbf{x}}_q, 
    \mathbf{x}^{\textrm{RV}}_{\mathcal{T}} \oplus \Tilde{\mathbf{x}}^{\textrm{RV}}_{q},
    \Tilde{\mathbf{y}}_{\mathcal{T}}^{\textrm{3D}} \oplus \Tilde{\mathbf{y}}_q),\\
    \mathbf{M}^{\textrm{3D}} &= \Psi(\mathbf{x}^{\textrm{3D}}, \mathbf{x}^{\textrm{RV}}), \mathbf{M}^{\textrm{3D}}\in\mathbb{R}^{(N+M)},\\
    (\mathbf{x}^{\textrm{3D}}_p, \Tilde{\mathbf{y}}^{\textrm{3D}}_p) &=
    (\mathbf{M}^{\textrm{3D}} \odot \mathbf{x}^{\textrm{3D}},
    \mathbf{M}^{\textrm{3D}} \odot \Tilde{\mathbf{y}^{\textrm{3D}}}), 
\end{align}
\endgroup
where $\oplus$ denotes the concatenation along the point axis and $\Tilde{\mathbf{y}}_{\mathcal{T}}^{\textrm{3D}}$ denotes the pseudo-labels of the raw scan. $\Psi$ represents the RV-based filtering operation, which preserves only one 3D point in $\mathbf{x}^{\textrm{3D}}$ for each pixel in $\mathbf{x}^{\textrm{RV}}_p$ and generates a valid mask $\mathbf{M}^{\textrm{3D}}$. $\odot$ denotes the point-wise selection based on the valid mask. The generated $(\mathbf{x}^{\textrm{3D}}_p, \Tilde{\mathbf{y}}^{\textrm{3D}}_p)$ is then utilized as the additional input pair for the subsequent 3D training process.

\subsection{SAM Mask Consistency (2D)}\label{Sec:Method_SAM}
For ROs with relatively small LiDAR cross sections as in Fig.~\ref{Fig:Intro}(b), the sparse supervision signals provided by the 3D point cloud cannot fully leverage the advantage of dense 2D input. To address this issue, we leverage the semantic mask from Segment Anything Model (SAM)~\cite{kirillov2023segment} as 2D prior knowledge to regularize the dense 2D predictions. As a powerful fundamental model with zero-shot generalization, SAM is able to provide semantic masks for unseen images without re-training as shown in Fig.~\ref{Fig:Main_Method}, albeit with unknown semantic categories. To leverage 2D prior masks from SAM, we propose a SAM mask consistency loss to encourage consistent pixel-wise predictions within each mask. Meanwhile, the entropy of the mask-wise mean prediction is minimized to encourage mask-wise confident predictions. Given a set of SAM masks $\mathbf{M^{\textrm{2D}}}=\{\mathbf{m}_i^{\textrm{2D}}\in\mathbb{R}^{H\times W}\}_{i=1}^{|\mathbf{M^{\textrm{2D}}}|}$ and pixel-wise 2D semantic predictions $\Tilde{\mathbf{p}}^{\textrm{2D}}\in\mathbb{R}^{H\times W\times C}$ before the pixel-to-point projection, our SAM consistency loss is defined as:
\begin{align}
    \mathcal{L}_{\textrm{SC}}^{\mathcal{T}} = \frac{1}{|\mathbf{M^{\textrm{2D}}}|}\sum^{|\mathbf{M^{\textrm{2D}}}|}_{i=1}(\textrm{MSE}(\Tilde{\mathbf{p}}^\textrm{2D}_{\mathbf{m}_i}) + \sum_{C}\Bar{\Tilde{\mathbf{p}}}^\textrm{2D}_{\mathbf{m}_i}\log{\Bar{\Tilde{\mathbf{p}}}^\textrm{2D}_{\mathbf{m}_i}}),
\end{align}
where $\Tilde{\mathbf{p}}^\textrm{2D}_{\mathbf{m}_i} = \mathbf{m}_i^\textrm{2D} \odot \Tilde{\mathbf{p}}^{\textrm{2D}}$ are the pixels located in the mask $\mathbf{m}_i^\textrm{2D}$ while $\textrm{MSE}(\cdot)$ denotes the mean-square-error computation. $\Bar{\Tilde{\mathbf{p}}}^\textrm{2D}_{\mathbf{m}_i}$ is the mean predictions of pixels in the mask, which is regularized by the Shannon entropy minimization. In practice, $\mathcal{L}_{\textrm{SC}}$ is computed for each image in the batch level and summarized before optimization.

\begin{table*}[t]
    \vspace{5pt}
    \centering
    \tiny
    \caption{Performance (mIoU) of SS2MM frameworks. Methods with $\dag$ are re-implementation based on their official code. The best performance is in bold while the second best is underlined.}
    \setlength{\tabcolsep}{4pt}
    \resizebox{.75\textwidth}{!}{
    \begin{tabular}{l l c c c c c c c c c c c c c c c c}
    \hline
    \multirow{2}{*}{\parbox{1cm}{\centering Modality}} & \multirow{2}{*}{\parbox{1.2cm}{Method}} & \multirow{2}{*}{Code} & \multirow{2}{*}{{$\Delta\textrm{Params}$}} & & \multicolumn{3}{c}{\centering USA$\rightarrow$Singapore} & & \multicolumn{3}{c}{\centering Day$\rightarrow$Night} & &  \multicolumn{3}{c}{\centering A2D2$\rightarrow$KITTI}\\
    \cline{6-8} \cline{10-12} \cline{14-16}
    & & & & & 2D & 3D & xM & & 2D & 3D & xM & & 2D & 3D & xM\\
    \hline
    \multirow{1}{*}{\parbox{1cm}{\centering -}} & Source only 
    & - & +0.0 & & 53.4 & 46.5 & 61.3 & & 42.2 & 41.2 & 47.8 & & 36.0 & 36.6 & 41.8\\
    \hline
    \multirow{4}{*}{\parbox{1cm}{\centering Uni-Modal}} 
    & Deep logCORAL~\cite{morerio2017minimal}
    & \cmark & +2.8M & & 52.6 & 47.1 & 59.1 & & 41.4 & 42.8 & 51.8 & & 35.8 & 39.3 & 40.3\\
    & MinEnt~\cite{vu2019advent}
    & \cmark & +28.3M & & 53.4 & 47.0 & 59.7 & & 44.9 & 43.5 & 51.3 & & 38.8 & 38.0 & 42.7\\
    & PL~\cite{li2019bidirectional}
    & \cmark & +0.0 & & 55.5 & 51.8 & 61.5 & & 43.7 & 45.1 & 48.6 & & 37.4 & 44.8 & 47.7\\
    & CLAN~\cite{luo2019taking}
    & \cmark & +2.8M & & 57.8 & 51.2 & 62.5 & & 45.6 & 43.7 & 49.2 & & 39.2 & 44.7 & 44.5\\
    \hline
    \multirow{10}{*}{\parbox{1cm}{\centering Multi-Modal}} 
    & $\textrm{xMUDA}^\dag$~\cite{jaritz2020xmuda}
    & \cmark & +0.8K & & 58.5 & 51.2 & 61.0 & & 47.7 & 45.1 & 50.1 & & 42.8 & 45.4 & 48.1\\ 
    & $\textrm{xMUDA+PL}^\dag$~\cite{jaritz2020xmuda}
    & \cmark & +0.8K & & 60.0 & 51.9 & 62.3 & & 47.9 & 45.8 & 50.9 & & 45.4 & 50.1 & 50.4\\
    & $\textrm{DsCML}^\dag$~\cite{peng2021sparse}
    & \cmark & +0.6M & & 39.0 & 38.8 & 45.2 & & 44.7 & 36.3 & 49.7 & & 26.3 & 37.0 & 35.6\\
    & $\textrm{DsCML+PL}^\dag$~\cite{peng2021sparse}
    & \cmark & +0.6M & & 39.8 & 41.7 & 46.9 & & 46.3 & 34.7 & 46.7 & & 26.6 & 37.2 & 36.0\\
    & AUDA~\cite{liu2021adversarial}
    & \xmark & +0.2M & & 59.8 & 52.0 & 62.7 & & 46.2 & 44.2 & 50.0 & & 38.3 & 46.0 & 44.0\\
    & AUDA+PL~\cite{liu2021adversarial}
    & \xmark & +0.2M & & 61.9 & 54.8 & \underline{65.6} & & 50.3 & \underline{49.7} & 52.6 & & 46.8 & 48.1 & 50.6\\
    & CMCL~\cite{xing2023cross}
    & \xmark & +34.1K & & \underline{62.0} & 54.2 & 64.5 & & 49.0 & 46.6 & 51.6 & & 42.8 & 47.7 & 48.0\\
    & CMCL+PL~\cite{xing2023cross}
    & \xmark & +34.1K & & \textbf{63.3} & \underline{57.1} & \textbf{66.7} & & 50.6 & \textbf{49.9} & \textbf{54.5} & & 47.6 & 51.3 & 52.8\\
    \cdashline{2-13}
    & MoPA+PL (Ours)
    & \cmark & +0.8K & & 60.6 & 57.0 & 63.5 & & \underline{51.4} & 47.5 & 53.4 & & \underline{48.8} & \underline{53.4} & \underline{53.5}\\
    & MoPA+PLx2 (Ours)
    & \cmark & +0.8K & & 61.7 & \textbf{57.2} & 64.1 & & \textbf{52.2} & 48.8 & \underline{54.2} & & \textbf{49.3} & \textbf{55.1} & \textbf{54.3}\\
    \hline
    \end{tabular}
    }
    \smallskip
    \label{Table:Overall_Comparison}
    \vspace{-10pt}
\end{table*}

\subsection{Overall Pipeline}\label{Sec:Method_Overall}
The overall pipeline of MoPA is illustrated in Fig.~\ref{Fig:Main_Method}. While MoPA can be easily integrated with all existing MM-UDA methods, we leverage xMUDA~\cite{jaritz2020xmuda} as our baseline due to its reproducibility. Specifically, when training on the source domain sample $(\mathbf{x}_{\mathcal{S}}^{\textrm{2D}}, \mathbf{x}_{\mathcal{S}}^{\textrm{3D}}, \mathbf{y}_{\mathcal{S}})$, our optimization objective follows xMUDA~\cite{jaritz2020xmuda}, formulated as:
\begin{align}
    \mathcal{L}^{\mathcal{S}}_{\textrm{CE}} &= \mathcal{F}(\mathbf{p}^\textrm{2D}_{\mathcal{S}}, \mathbf{y}_{\mathcal{S}}) +
    \mathcal{F}(\mathbf{p}^\textrm{3D}_{\mathcal{S}}, \mathbf{y}_{\mathcal{S}}),\label{Eq:Source_CE}\\
    \mathcal{L}^{\mathcal{S}}_{\textrm{xM}} &= D_{KL}(\mathbf{p}^\textrm{3D}_{\mathcal{S}}\|\Acute{\mathbf{p}}^{\textrm{2D}}_{\mathcal{S}}) + 
    D_{KL}(\mathbf{p}^\textrm{2D}_{\mathcal{S}}\|\Acute{\mathbf{p}}^{\textrm{3D}}_{\mathcal{S}}),\label{Eq:Source_xM}
\end{align}
where $\mathcal{F}(\cdot)$ denotes the point-wise cross-entropy function. $\Acute{\mathbf{p}}^{\textrm{3D}}_{\mathcal{S}}$ and $\Acute{\mathbf{p}}^{\textrm{2D}}_{\mathcal{S}}$ denote the prediction from the auxiliary 3D and 2D classifiers as in xMUDA, respectively. While training on the target domain, the optimization objectives of xMUDA are also preserved, including $\mathcal{L}^{\mathcal{T}}_{\textrm{CE}}$ and $\mathcal{L}^{\mathcal{T}}_{\textrm{xM}}$ similarly computed as Eq.~\ref{Eq:Source_CE} and~\ref{Eq:Source_xM} with $\mathbf{y}_\mathcal{S}$ replaced by pseudo-labels $\Tilde{\mathbf{y}}_{\mathcal{T}}$. Further, given the object-inserted sample $(\mathbf{x}_p^{\textrm{3D}}, \Tilde{\mathbf{y}}_p^{\textrm{3D}})$ and the SAM consistency loss, the overall loss $\mathcal{L}$ is formulated as:
\begingroup
\allowdisplaybreaks
\begin{align}
    % \mathcal{L}^{\mathcal{T}}_{\textrm{V-CE}} &= \frac{1}{N} 
    % \sum_{n=1}^{N}\sum_{c=1}^{C}{
    % \phi_\theta^\textrm{3D}(\mathbf{x}^{\textrm{3D}}_p)_{(n,c)}\log{\Tilde{\mathbf{y}}^{\textrm{3D}}_{p,(n,c)}}},\label{Eq:Source_CE}\\
     \mathcal{L}^{\mathcal{T}}_{\textrm{V-CE}} &= \mathcal{F}(\phi_\theta^\textrm{3D}(\mathbf{x}^{\textrm{3D}}_p), \Tilde{\mathbf{y}}^{\textrm{3D}}_{p}),\label{Eq:Target_CE}\\
    \mathcal{L} &= \underbrace{\mathcal{L}^{\mathcal{S}}_{\textrm{CE}} +
    \lambda_{\textrm{xM}}^{\mathcal{S}}\mathcal{L}^{\mathcal{S}}_{\textrm{xM}} +
    \mathcal{L}^{\mathcal{T}}_{\textrm{CE}} +
    \lambda_{\textrm{xM}}^{\mathcal{T}}\mathcal{L}^{\mathcal{T}}_{\textrm{xM}}}_\text{xMUDA} \nonumber \\
    & + \underbrace{\lambda_{\textrm{V-CE}}^{\mathcal{T}}\mathcal{L}^{\mathcal{T}}_{\textrm{V-CE}} +
    \lambda_{\textrm{SC}}^{\mathcal{T}}\mathcal{L}^{\mathcal{T}}_{\textrm{SC}}}_\text{MoPA}, \label{Eq:Total_Loss}
\end{align}
\endgroup
where $\lambda_{\textrm{xM}}^{\mathcal{S}}, \lambda_{\textrm{xM}}^{\mathcal{T}}, \lambda_{\textrm{V-CE}}^{\mathcal{T}}, \lambda_{\textrm{SC}}^{\mathcal{T}}$ are the pre-defined coefficient for each loss components.

Considering the inferior RO segmentation of offline pseudo-labels, an exponential moving average (EMA) is introduced, which utilizes a slowly updated teacher to generate the pseudo-labels in an online manner. When enabling EMA, the teacher network $\phi^{m}_{\theta'}$ is initialized from the student network $\phi^{m}_{\theta}$ and updated as the moving average of the student one for each iteration ($\theta'_{i}=\alpha\theta'_{i-1} + \theta_{i}$ where $i$ indicate the training iteration). To facilitate cross-modal learning, the pseudo-labels from EMA for each modality are occasionally swapped to the cross-modal one $\Tilde{\mathbf{y}}_\mathcal{T}^\textrm{xM}$ derived from $\mathbf{p}_\mathcal{T}^\textrm{xM}$ (i.e., $\Tilde{\mathbf{y}}_\mathcal{T}^m=\Tilde{\mathbf{y}}_\mathcal{T}^\textrm{xM}$ at a percentage of $p_{\textrm{xM}}$).

% given the object-inserted sample $(\Tilde{\mathbf{x}}_\textrm{all}^{\textrm{3D}}, \Tilde{\mathbf{y}}_\textrm{all}^{\textrm{3D}})$ generated by our VGI
% \frac{1}{N} 
%     \sum_{n=1}^{N}\sum_{c=1}^{C}{
%     \mathbf{p}^m_{\mathcal{S},(n,c)}\log{\mathbf{y}_{\mathcal{S},(n,c)}}}

\section{Experimental Results}
\subsection{Benchmarks and Settings}
\noindent \textbf{Benchmarks:} following xMUDA~\cite{jaritz2020xmuda}, we evaluate our MoPA on 3 MM-UDA benchmarks: (i) USA-to-Singapore (U-to-S), (ii) Day-to-Night (D-to-N), and (iii) A2D2-to-SemanticKITTI (A-to-S). For U-to-S and D-to-N, nuScenes~\cite{caesar2020nuscenes} is leveraged as their data source. Specifically, U-to-S differs in layout and infrastructure between the source and target domains while D-to-N contains images recorded in different illumination levels. Following prior works~\cite{jaritz2020xmuda}, points within each bounding box are assigned with their corresponding object label, where the others are labeled as background. For A-to-S, A2D2~\cite{geyer2020a2d2} and SemanticKITTI~\cite{behley2019semantickitti} are utilized as the source and the target domains, where the domain discrepancy lies in the pattern and density differences of 3D point clouds. For a fair comparison, we strictly follow the organization of all benchmarks in prior MM-UDA methods~\cite{jaritz2020xmuda,peng2021sparse}. We utilize mean Intersection over Union (mIoU) as our evaluation metric same as~\cite{jaritz2020xmuda, peng2021sparse}.

\noindent \textbf{Implementation Details:} following previous works~\cite{jaritz2020xmuda,peng2021sparse,li2022cross}, ResNet34~\cite{he2016deep} and U-Net-based SparseConvNet~\cite{graham20183d} are leveraged as our 2D and 3D backbones. Since we utilize xMUDA~\cite{jaritz2020xmuda} as our baseline, we strictly follow most hyper-parameters settings as xMUDA, including the optimizer, learning scheduler, and xMUDA-related coefficient settings. For VGI, we collect our prior objects from Waymo dataset~\cite{sun2020scalability} thanks to its data variety. Specifically, ``Pedestrian'', ``Bicycle'', and ``Motorcycle'' are the prior objects of interest for all benchmarks. Each class of interest in the object pool contains 1000 object instances at most and we randomly insert 1 prior object instance for each raw input scan, where a voxel size of 50 cm is used for overlap checking and grounding. For SAM masks, the ViT-h variant is utilized to generate 2D prior masks in an offline manner, where masks larger than 1/10 image size are filtered out for mask refinement. For MoPA coefficients, we empirically set $\lambda_\textrm{V-CE}^\mathcal{T}$ and $\lambda_{\textrm{SC}}^{\mathcal{T}}$ as 0.1 and 0.01. The online pseudo-labels are enabled at the 70k iteration (100k in total) and $p_\textrm{xM}$ is set to 0.7. All experiments are conducted with PyTorch~\cite{paszke2019pytorch} on a single NVIDIA RTX 3090.

\begin{figure}[t]
    \centering
    \includegraphics[width=.73\linewidth]{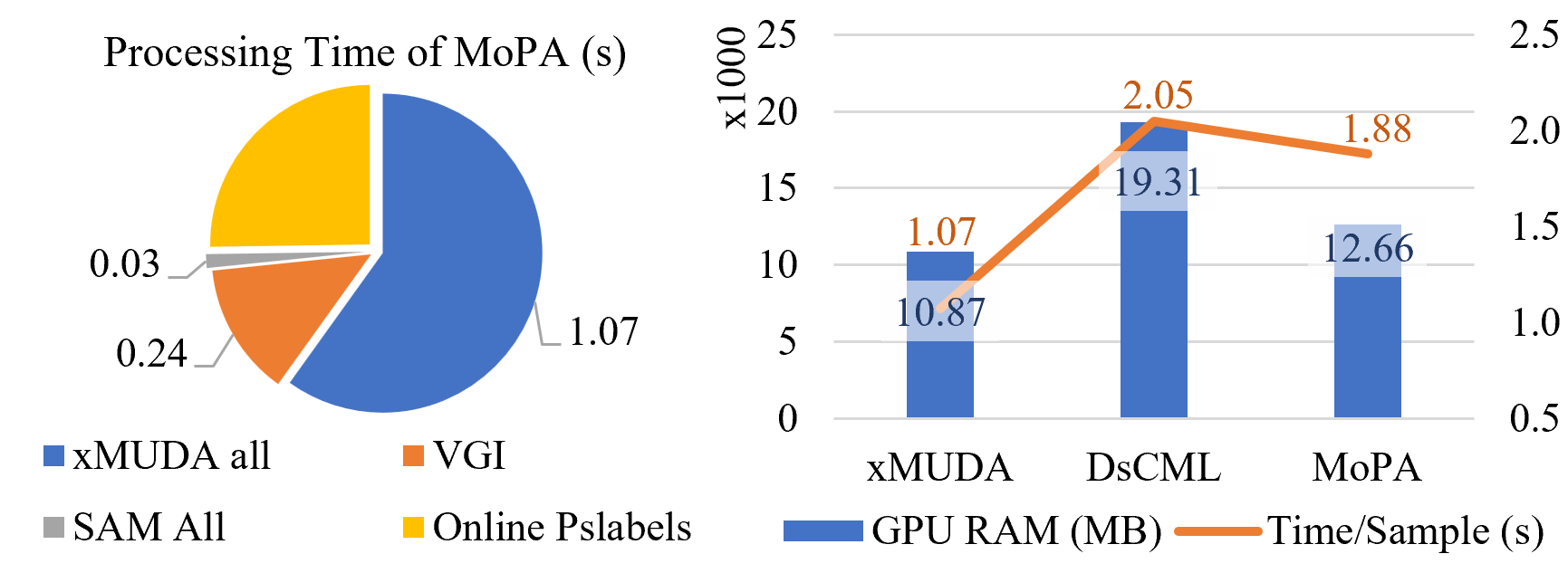}
    \vspace*{-6pt}
    \caption{Processing time of MoPA per batch (left) and efficiency comparison with xMUDA and DsCML (right). Postfix ``+PL'' is omitted for clarity.}
    \label{Fig:Efficiency_Test}
    \vspace{-5pt}
\end{figure}

\begin{figure*}[t]
    \centering
    \vspace{5pt}
    \includegraphics[width=.8\textwidth]{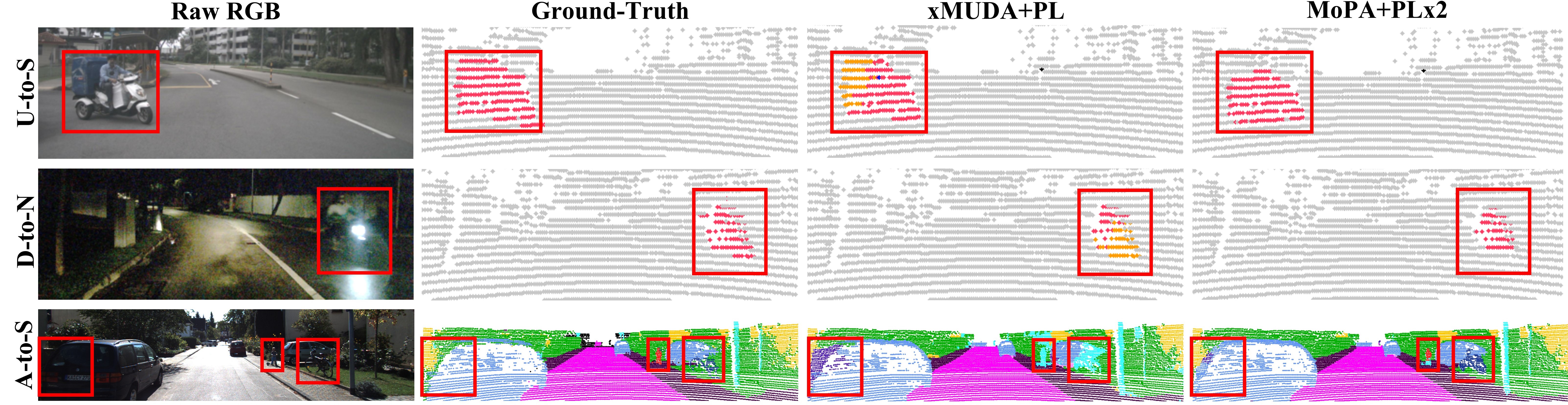}
    \vspace*{-6pt}
    \caption{Qualitative results of cross-modal predictions from MoPA+PLx2 and xMUDA+PL. Here different colors represent different semantic classes. Our MoPA+PLx2 obviously outperforms xMUDA+PL when segmenting the challenging RO. }
    \label{Fig:Ablation_Viz}
    \vspace{-10pt}
\end{figure*}

\subsection{Comparison with SOTA Methods}
\noindent \textbf{Overall Performance.}
Table.~\ref{Table:Overall_Comparison} shows the performance comparison with previous MM-UDA methods. Leveraging the pseudo-labels generated by xMUDA, MoPA+PL achieves a noticeable improvement compared to the baseline xMUDA+PL on U-to-S and D-to-N. On A-to-S, this improvement significantly expands to more than $3\%$ on all predictions on all single-modal (i.e., 2D or 3D) and cross-modal predictions (i.e., xM). Considering the biased pseudo-labels generated by xMUDA, we introduce the additional $\textrm{MoPA+PL}\times2$ which leverages the pseudo-labels generated by MoPA+PL instead. This enables our $\textrm{MoPA+PL}\times2$ to achieve competitive performance on U-to-S and D-to-N compared to the previous SOTA method CMCL+PL~\cite{xing2023cross}, with a minor relative drop of less than $2.4\%$ but introducing $97.7\%$ fewer trainable parameters. In terms of the more challenging A-to-S with double classes as others, our $\textrm{MoPA+PL}\times2$ outperforms all previous methods by more than $1.5\%$ on the cross-modal prediction. Please note that \textit{we report the re-implementation results of both xMUDA~\cite{jaritz2020xmuda} and DsCML~\cite{peng2021sparse} by their official code}.

\begin{table}[t]
    \centering
    \tiny
    \caption{RO performance comparison on A-to-S.}
    \vspace{-5pt}
    \setlength{\tabcolsep}{4pt}
    \resizebox{.75\linewidth}{!}{
    \begin{tabular}{l c c c c c c c}
    \hline
    \multirow{2}{*}{\parbox{1.2cm}{Method}} & \multicolumn{3}{c}{\centering Person} & & \multicolumn{3}{c}{\centering Bike}\\
    \cline{2-4} \cline{6-8} 
    & 2D & 3D & xM & & 2D & 3D & xM\\
    \hline
    xMUDA & 13.0 & 25.4 & 21.8 & & 24.4 & 42.9 & 36.2\\
    % xMUDA+PL & 25.2 (+12.2) & 29.8 (+4.4) & 34.0 (+12.2) & & 32.7 (+8.3) & 47.1 (+4.2) & 42.9 (+6.7)\\
    % xMUDA+PL & +12.2 & +4.4 & +12.2 & & +8.3 & +4.2 & +6.7\\
    xMUDA+PL & 25.2 & 29.8 & 34.0 & & 32.7 & 47.1 & 42.9\\
    \hdashline
    % MoPA+PL & 26.3 & 33.6 & 41.1 (+7.1) & & 40.9 & 53.1 & 54.0 (+7.1)\\
    % MoPA+PLx2 & 42.2 & 44.3 & 53.8 (+19.8) & & 44.4 & 57.6 & 55.2 (+12.3)\\
    MoPA+PL & 26.3 & 33.6 & 41.1 & & 40.9 & 53.1 & 54.0\\
    MoPA+PLx2 & \textbf{42.2} & \textbf{44.3} & \textbf{53.8} & & \textbf{44.4} & \textbf{57.6} & \textbf{55.2}\\
    \hline
    \end{tabular}}
    \smallskip
    \label{Table:Ablation_RO}
    \vspace{-5pt}

    \centering
    \caption{Ablation study of MoPA Components.}
    \vspace{-5pt}
    \resizebox{.7\linewidth}{!}{
    \begin{tabular}{c|c c c| c c c}
    \hline
    Method & EMA & SAM & VGI & 2D & 3D & xM \\
    \hline
    xMUDA+PL  & - & - & - & 45.4 & 50.1 & 50.4 \\
    \hline
    \multirow{7}{*}{\centering MoPA+PL}
    & \cmark &        &        & 42.7 & 51.6 & 47.8\\
    &        & \cmark &        & 45.6 & 52.1 & 51.3\\
    &        &        & \cmark & 45.3 & 51.3 & 51.1\\
    & \cmark & \cmark &        & 43.7 & 50.5 & 49.6\\
    & \cmark &        & \cmark & 45.2 & 51.9 & 51.5\\
    &        & \cmark & \cmark & 46.7 & 52.4 & 52.7\\
    \cline{2-7}
    & \cmark & \cmark & \cmark & \textbf{48.8} & \textbf{53.4} & \textbf{53.5}\\
    \hline
    \end{tabular}
    }
    \label{Table:MoPA_Component}
    \vspace{-6pt}
\end{table}

\noindent \textbf{Training Efficiency.} To justify the efficiency of MoPA, we present the processing time and GPU RAM consumption of MoPA on A-to-S in Fig.~\ref{Fig:Efficiency_Test}. Specifically, despite the utilization of grid-search-based validity checking, thanks to voxelization and GPU acceleration, VGI and the addition network forward process lead to a minor 0.24s of time overhead where the VGI itself only takes about 0.1s. The main computational overhead comes from the online pseudo-labels with EMA, while we only enable it for the last $30\%$ of the training process. Meanwhile, MoPA is more efficient in terms of GPU consumption compared to the existing method DsCML~\cite{peng2021sparse}, which consumes $33.5\%$ less GPU RAM yet achieves more than $35.8\%$ improvement.

\subsection{Ablation Studies and Analysis}
\noindent \textbf{Effect of MoPA on ROs.} MoPA is designed to improve RO performance and we justify this by comparing MoPA with xMUDA and xMUDA+PL in Table.~\ref{Table:Ablation_RO}. While both leveraging the pseudo-labels generated by xMUDA, our MoPA+PL noticeably outperforms xMUDA+PL on both ROs of our interest (note that ``Motorcycle'' is merged in ``Bicycle'' in A-to-S) on all single-modal or multi-modal predictions. Utilizing the rectified pseudo-labels from MoPA+PL, MoPA+PLx2 further expands this performance gap to more than $10\%$ on all outputs for both classes of interest, which justifies the effectiveness of MoPA in improving RO performance.

\noindent \textbf{Effect of Each Component of MoPA.} Table.~\ref{Table:MoPA_Component} illustrates our validation of each MoPA component on A-to-S. Specifically, utilizing only SAM or VGI results in an improvement of $0.9\%$ and $0.7\%$, respectively, and their combination further enlarges the gap to $2.3\%$. Interestingly, utilizing only EMA update or EMA update with SAM results in an inferior performance compared to the baseline ($-2.6$ and $-0.8\%$). This is mainly because utilizing the EMA update without effective regularization on pseudo-labels would result in a more biased performance.  With VGI which regularizes pseudo-labels by inserting RO objects, the utilization of EMA update boosts the performance ($+0.4\%$ on pure VGI and $+0.8\%$ on VGI+SAM).

\begin{figure}[t]
    \centering
    \includegraphics[width=.74\linewidth]{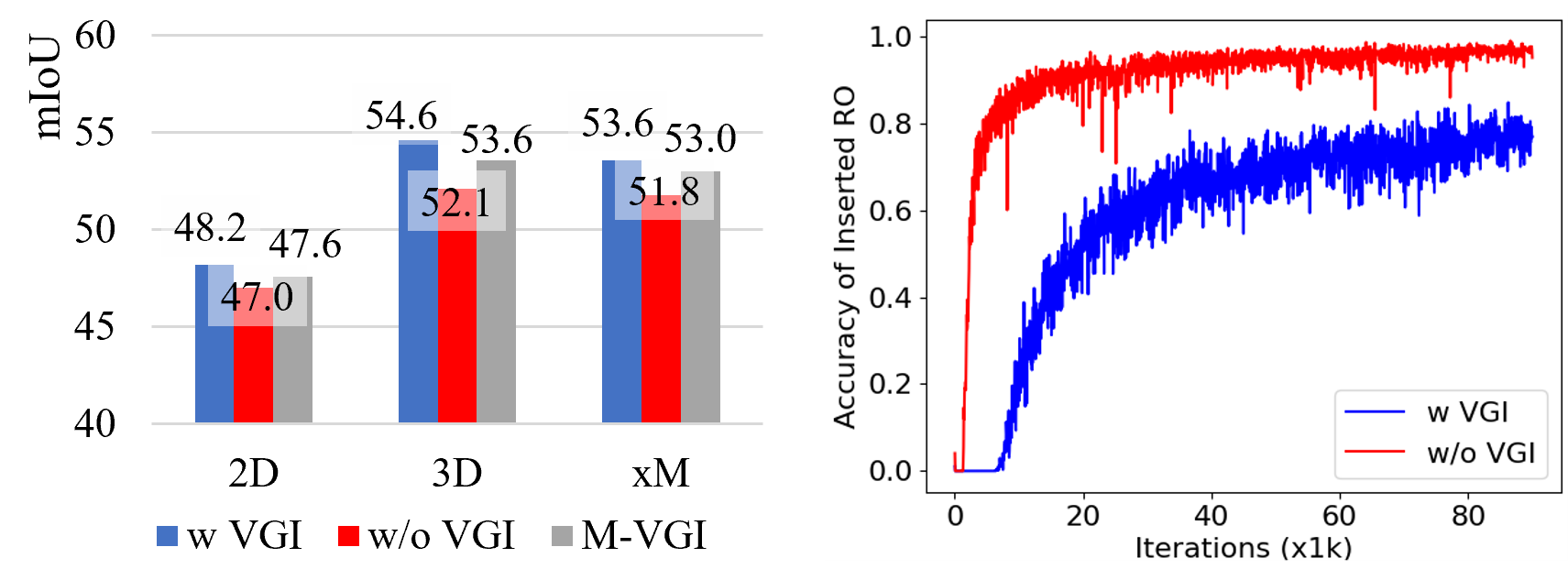}
    \vspace*{-6pt}
    \caption{Comparison of insertion with VGI, without VGI, and with M-VGI (left) and the inserted RO training accuracy with or without VGI (right).}
    \label{Fig:Ablation_VGI}
    \vspace{-6pt}
\end{figure}

\noindent \textbf{RO Insertion through VGI.}
Fig.~\ref{Fig:Ablation_VGI} presents the comparison among insertion with VGI (w VGI), without VGI (w/o VGI), and with VGI but multiple objects (M-VGI). Compared with w/o VGI, w VGI achieves a notable improvement of more than $1.2\%$ on all predictions. Based on the training accuracy of inserted RO from w VGI and w/o VGI, it can be seen that the insertion without VGI leads to more early saturation of recognizing inserted RO due to trivial solutions. This justifies our design of VGI to avoid introducing artificial artifacts during insertion. On the other hand, M-VGI might not be a better alternative for front-view-only cases compared to VGI since it further hampers the limited valid insertion locations, leading to less insertion variance for learning. 

\noindent \textbf{Qualitative results.}
Fig.~\ref{Fig:Ablation_Viz} illustrates the qualitative results of our MoPA+PLx2 compared with our baseline method xMUDA+PL. Specifically, we visualize the cross-modal predictions of our MoPA+PLx2 and xMUDA+PL on samples of three different benchmarks. For D-to-N and U-to-S where xMUDA+PL struggles to segment out the motorcycle, our MoPA+PLx2 generates more accurate predictions with fewer true-negatives. For the more challenging scan from A-to-S, our MoPA+PLx2 clearly outperforms xMUDA+PL in segmenting the pedestrian and bicycle located far from the sensor. This justifies the effectiveness of our MoPA, improving the performance on challenging RO.

\section{Conclusions}
In this paper, we present a novel method called MoPA that leverages multi-modal prior knowledge for MM-UDA. By introducing more prior ROs with accurate pseudo-labels and learning from foundation models, MoPA can effectively improve RO performance. Extensive experiments justify the effectiveness of MoPA. The utilization of MoPA can potentially improve the performance of classes of interest in a flexible manner, which can bridge the gap between performance benchmarking and real-world application. 

{\small
\bibliographystyle{IEEEtran}
\bibliography{icra24}
}
\end{document}